\documentclass[10pt,twocolumn,letterpaper]{article}

\usepackage{cvpr}              
\makeatletter
\@namedef{ver@everyshi.sty}{}
\makeatother
\usepackage{tikz}
\usepackage{graphicx}
\usepackage{amsmath}
\usepackage{amssymb}
\usepackage{booktabs}
\usepackage{algorithm2e}
\usepackage{algpseudocode}
\usepackage[accsupp]{axessibility}  

\usepackage[colorinlistoftodos,prependcaption,textsize=tiny]{todonotes}

\usepackage[pagebackref,breaklinks,colorlinks]{hyperref}
\usepackage{xspace}

\usepackage[capitalize]{cleveref}
\crefname{section}{Sec.}{Secs.}
\Crefname{section}{Section}{Sections}
\Crefname{table}{Table}{Tables}
\crefname{table}{Tab.}{Tabs.}

\newcommand{\name}{ProgDTD\xspace}

\begin{document}

\title{\name: Progressive Learned Image Compression\\ with Double-Tail-Drop Training}

\author{Ali Hojjat\textsuperscript{1},
Janek Haberer\textsuperscript{1}, 
Olaf Landsiedel\textsuperscript{1,2}
\\
\textsuperscript{1}Kiel University, Germany;
\textsuperscript{2}Chalmers University of Technology, Sweden 
\\
{\tt\small \{aho, jha, ol\}@informatik.uni-kiel.de}
}

\maketitle

\begin{abstract}

Progressive compression allows images to start loading as low-resolution versions, becoming clearer as more data is received.
This increases user experience when, for example, network connections are slow. 
Today, most approaches for image compression, both classical and learned ones, are designed to be non-progressive. This paper introduces \name, a training method that transforms learned, non-progressive image compression approaches into progressive ones. 
The design of \name is based on the observation that the information stored within the bottleneck of a compression model commonly varies in importance. 
To create a progressive compression model, \name modifies the training steps to enforce the model to store the data in the bottleneck sorted by priority. We achieve progressive compression by transmitting the data in order of its sorted index. 
\name is designed for CNN-based learned image compression models, does not need additional parameters, and has a customizable range of progressiveness. 
For evaluation, we apply \name to the hyperprior model, one of the most common structures in learned image compression. 
Our experimental results show that \name performs comparably to its non-progressive counterparts and other state-of-the-art progressive models in terms of MS-SSIM and accuracy.
\end{abstract}

\section{Introduction}
\label{sec:intro}

Image compression has been an active research field for many years and has led to numerous classical compression methods such as JPEG \cite{wallace1992jpeg}, JPEG 2000\cite{skodras2001jpeg}, WebP\cite{webp} and BPG\cite{bpg}. 
The rise of deep learning\cite{krizhevsky2017imagenet} inspired new methods that employ the neural networks' power for learned image compression \cite{lee2018context, gregor2016towards, toderici2017full, toderici2015variable}. 
Among these, particularly the recent success of the variational image compression led to new, state-of-the-art methods, which often perform on par or even better than established, classical and deep methods \cite{minnen2018joint, cheng2020learned, su2020scalable, balle2018variational, Balle2016end}. 

Most image compression methods, both classic such as original JPEG  \cite{wallace1992jpeg}, Webp \cite{webp}, BPG\cite{bpg} and learned ones \cite{minnen2018joint, cheng2020learned, balle2018variational, Balle2016end}, are non-progressive. 
Thus, these expect the complete compressed image to be available for decoding. Such availability of an entire file is, however, a challenge in many settings: for example, slow network connections often delay the transmissions. As a result, a user or system experiences a delay until the image can be reconstructed for viewing or further processing. Progressive compression \cite{ohm2005advances} addresses this problem, and the decoder can obtain an initial preview even with a small portion of the data. Later, by receiving the rest of the bits, the decoder can reconstruct a better-quality image.
However, most learned approaches to image compression are non-progressive, and only a few are progressive \cite{lee2022dpict, toderici2017full, johnston2018improved, cai2018efficient, diao2020drasic}.

\begin{figure}[tb]
  \centering
   \includegraphics[trim=10 10 15 10 10, width=.45\textwidth]{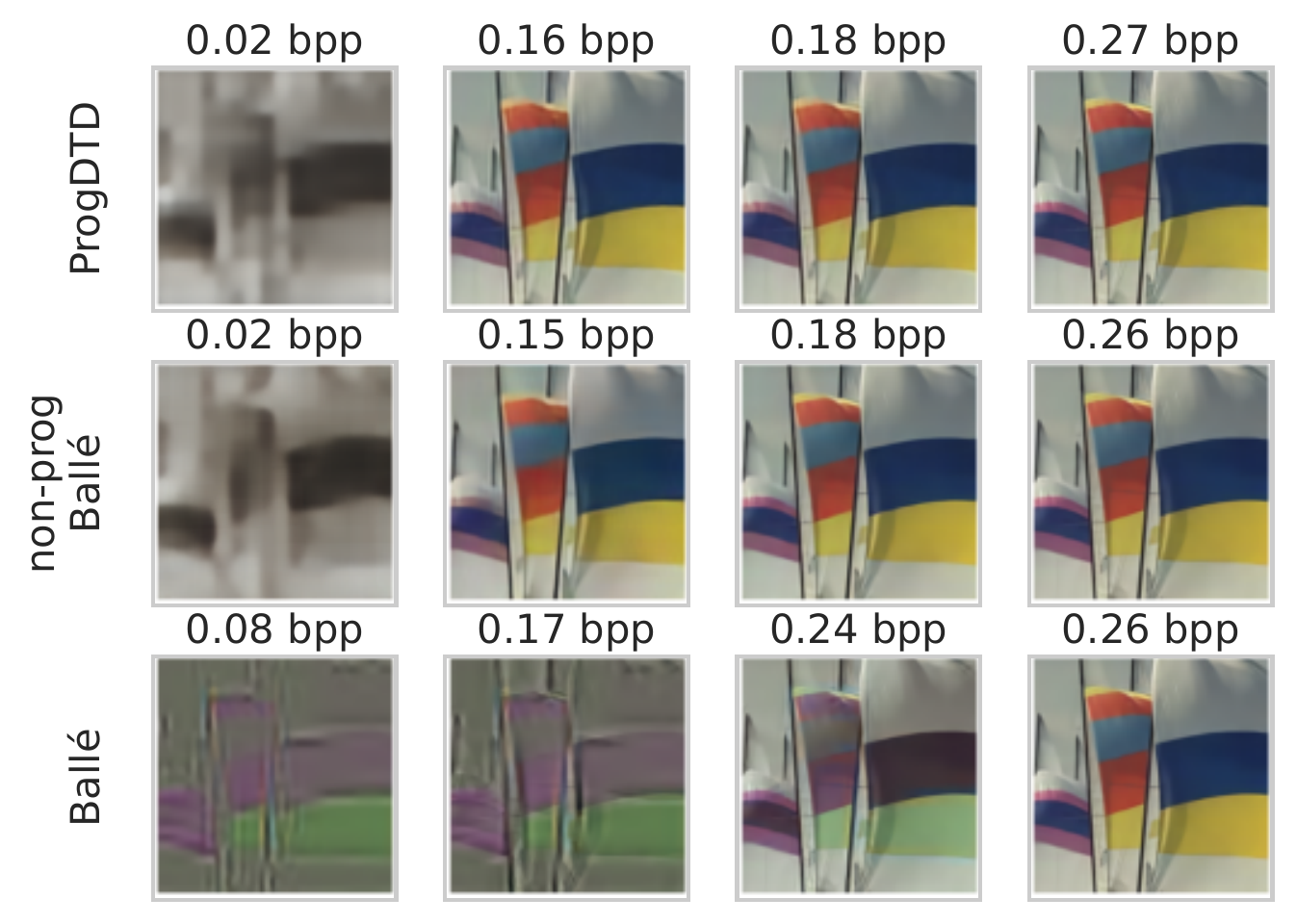}
   \caption{Qualitative comparison of reconstructed images from: \emph{\name} (trained with $\lambda=0.01$), \emph{non-progressive-Ball\'{e}} (trained with $\lambda=0.0001, 0.001,0.005, 0.01$) and \emph{standard-Ball\'{e}} (trained with $\lambda=0.01$). \emph{\name} and \emph{non-progressive-Ball\'{e}} reconstruct images of similar quality; meanwhile, removing only a few bits in \emph{standard-Ball\'{e}} leads to a significant degradation in quality.}
   \label{fig:Plots}
\end{figure}

In this paper, we introduce \name, a method to transform learned, non-progressive image compression approaches into progressive ones. Further, to showcase its efficiency, we apply \name to the Ball\'{e} \cite{balle2018variational} architecture, which is a widely adopted, state-of-the-art learned image compression model, and acts as a base architecture for many recent models \cite{choi2019variable, minnen2018joint}.
The design of \name builds on the following observation: 
Previous research \cite{koike2020stochastic, scholz2008nonlinear, wold1987principal} shows that the information stored within the bottleneck of a compression model commonly varies in terms of importance, i.e., its availability impacts the reconstructed image differently. 

To create a progressive compression model, we identify the most important data for each input within the bottleneck and transmit these ordered by their significance. For this, we modify the training steps of the model to enforce it to store the data in the bottleneck sorted by priority. 
We build on the tail-drop technique \cite{koike2020stochastic} and, as an example, apply it to the training of both the latent and hyper latent bottlenecks of the Ball\'{e} model \cite{balle2018variational}. We call this process double-tail-drop (DTD).

Our experimental results show that our progressive compression model trained with double-tail-drop has comparable performance in terms of MS-SSIM and accuracy and has a slight drop in PSNR compared to \emph{non-progressive-Ball\'{e}} (ensemble of Ball\'{e} models trained for different $\lambda$) and other state-of-the-art models.  A key benefit of our progressive model is that we do not add any parameters to the model. Instead, we modify the training method to sort the information in the bottleneck in order of importance. Also, \name has a customizable range of progressiveness, so we can choose the desirable bitrate range.
Figure \ref{fig:Plots} shows a qualitative comparison of \name, \emph{non-progressive-Ball\'{e}} and \emph{standard-Ball\'{e}}.
It underlines that \name and \emph{non-progressive-Ball\'{e}} achieve similar results, whereas the \emph{standard-Ball\'{e}} model requires the whole latent for a meaningful reconstruction of an image. 
Overall, the contributions of \name are as follows:
\begin{enumerate}
\itemsep0em
\item We introduce \name, a progressive image compression method that enables a non-progressive model to become progressive.
\item \name is a training approach that does not need additional parameters and is designed for learned image compression.
\item \name has a customizable range of progressiveness.
\item \name performs on par compared to its non-progressive version and other state-of-the-art benchmarks.
\end{enumerate}

The remainder of this paper is structured as follows: In Section \ref{sec: Related Work}, we present related works. Then, in Section \ref{sec: Proposed Algorithm}, we introduce the design of \name and its training.
In Section \ref{sec:result}, we evaluate \name by comparing it to the state-of-the-art. 
Section \ref{sec:conclusion} concludes the paper.


\section{Related Work}
\label{sec: Related Work}
In this section, we introduce the required related work on learned image compressions, learned variable bitrate compression, and deep progressive image compression.

\subsection{Learned image compression}

Most of the learned image compression methods are based on autoencoders or an extension of them. In these types of networks, an encoder \(g_a(x; \phi_g)\) compresses the input \(x\) into a latent space $y$. In deep image compression, authors usually call this the $Analysis Network$. 
Next, entropy models like Huffman encoding or arithmetic encoding \cite{langdon1981compression} compress and quantize the latent space $y$ to $\hat{y}$. 
During training, random noise often simulates quantization loss and helps the model to adapt to quantization effects \cite{Balle2016end}. 
For decoding, a so called $Synthesis Network$, receives the bits, recovers $\hat{y}$ and another network $g_s(\hat{y}; \theta_g)$ reconstructs $\hat{x}$. 

Ball\'{e} \etal \cite{balle2018variational} introduce a hyperprior for extracting spatial correlation in images. 
It assumes that correlations in each image are normal distributions with a mean of zero, and the task of the hyperprior network is to predict the standard deviation and the location of these distributions:
\begin{equation}
  p_{\boldsymbol{\hat{y}}|\boldsymbol{\hat{z}}}(\boldsymbol{\hat{y}}|\boldsymbol{\hat{z}})   \sim   N(\boldsymbol{0},\boldsymbol{\sigma^2})
  \label{eq:1}
\end{equation}
Minnen \etal \cite{minnen2018joint} consider these correlations as $N(\boldsymbol{\mu},\boldsymbol{\sigma^2})$.
Cheng \etal \cite{cheng2020learned} extend this work and assume that these correlations contain multiple distributions and detect them with a GMM (Gaussian Mixture Model). 
In other work, Cui \etal \cite{cui2021asymmetric} use asymmetric normal distributions $N(\boldsymbol{\mu},\boldsymbol{\sigma^2_1}, \boldsymbol{\sigma^2_2})$ to simulate these correlations. Rippel and Bourdev \cite{rippel2017real}, Tschannen \etal \cite{tschannen2018deep}, and
Agustsson \etal \cite{agustsson2019generative} adapt GANs~\cite{goodfellow2020generative} for image compression, and Nakanishi \etal \cite{nakanishi2019neural} use multi-scale autoencoders to compress images at multiple levels.
Nonetheless, deep compression methods commonly compress images at one specific, fixed bitrate.

\subsection{Variable bitrate learned image compression}

Some compression methods provide variable bitrates due to adaptive models \cite{toderici2015variable} or by employing dynamic quantization \cite{theis2017lossy, choi2019variable}. 
For example, Yang \etal \cite{yang2020variable} propose a modulated network allowing the encoder and decoder architecture to provide variable bitrates. 
Cai \etal \cite{cai2018efficient} introduce a CNN-based multi-scale decomposition transformation with content-adaptive rate allocation for variable bitrates. 
Yang \etal \cite{yang2021slimmable} use swimmable networks to train one compression model for different bitrates: 
At low bitrates, they only use parts of the network. 

\subsection{Progressive learned image compression}

In the field of progressive, learned compression, many approaches employ RNNs.
For example, Toderici \etal \cite{toderici2015variable} introduce an iterative LSTM based \cite{hochreiter1997long} compression method. 
First, it feeds the image to the LSTM-based compression model, calculates the residual, and feeds it again. 
After T repetitions, this design results in a T-step progressive compression. 
In further work, Toderici \etal \cite{toderici2017full} improve upon this by utilizing a Pixel-RNN \cite{van2016pixel}.
Cai \etal \cite{cai2019novel} introduce a progressive compression method based on a two-level encoder-decoder. 
The first level encodes the input image into a basic representation with low quality, and the second level encodes the input image into a higher-quality enhancement representation. 
Gregor \etal \cite{gregor2016towards} introduce a GAN-based compression method with RNNs which reconstructs the input image by feeding compressed data to the generative models. 
The benefit of GANs is that they can reconstruct the image even with a small part of the data and produce a more accurate reconstruction by getting more and more data. 
Lee \etal \cite{lee2022dpict} proposes to create a progressive compression method based on trit-planes.

Most of these progressive methods are specifically designed neural networks for progressiveness and can be very complex. 
In this paper, we propose \name, which is a training method based on tail-drop~\cite{koike2020stochastic} (see Figure \ref{fig:Tail-Drop}). We adopt the tail-drop idea in \name to work with CNNs and demonstrate that with a hyperprior-based model. \name can then add progressiveness without adding any parameters or complexity and has a customizable range of progressiveness.

\section{Proposed Algorithm}
\label{sec: Proposed Algorithm}
In this section, we introduce \name, a method to transform learned, non-progressive image compression approaches into progressive ones. 
To demonstrate the effectiveness of our method, we integrate it into the Ball\'{e} \etal \cite{balle2018variational} architecture, which is a widely adopted, state-of-the-art learned image compression model, and acts as a base architecture for many recent models \cite{choi2019variable, minnen2018joint}.
Our approach is not restricted to this particular architecture and is designed for learned image compression models with multiple latent representations. 
In Section \ref{sec: Model architecture}, we briefly introduce and explain the Ball\'{e} model, then in Sections \ref{sec: Rateless autoencoder} and \ref{sec: Training procedure}, we explain how we design and integrate \name into the Ball\'{e} architecture.

\subsection{Model architecture} \label{sec: Model architecture}
As discussed in Section \ref{sec: Related Work}, we formulate image compression as:
\begin{align}
& \boldsymbol{y}=g_a(\boldsymbol{x} ; \boldsymbol{\phi}) \\
& \hat{\boldsymbol{y}}=Q(\boldsymbol{y}) \\
& \hat{\boldsymbol{x}}=g_s(\hat{\boldsymbol{y}} ; \boldsymbol{\theta})
\end{align}

where $\boldsymbol{x}, \hat{\boldsymbol{x}}, \boldsymbol{y}$, $\hat{\boldsymbol{y}}$ and $Q$ are raw images, reconstructed images, latent representation before quantization, latent representation after quantization and quantization respectively. Furthermore, $\phi$ and $\theta$ are neural network-based transformations consisting of convolution, de-convolution, GDN, and IGDN \cite{Balle2015density}. 
As common, we approximate quantization as a uniform noise $\mathcal{U}\left(-\frac{1}{2}, \frac{1}{2}\right)$ to generate noisy codes $\hat{\boldsymbol{y}}$ during training. Ball\'{e} \etal define a hyperprior, by introducing side information $\boldsymbol{z}$ to capture spatial dependencies among the elements of $\boldsymbol{y}$, formulated as:
\begin{align}
\boldsymbol{z} & =h_a\left(\boldsymbol{y} ; \boldsymbol{\phi}_{\boldsymbol{h}}\right) \\
\hat{\boldsymbol{z}} & =Q(\boldsymbol{z}) \\
p_{\hat{\boldsymbol{y}} \mid \hat{\boldsymbol{z}}}(\hat{\boldsymbol{y}} \mid \hat{\boldsymbol{z}}) & \leftarrow h_s\left(\hat{\boldsymbol{z}} ; \boldsymbol{\theta}_{\boldsymbol{h}}\right)
\end{align}

where $h_a$ and $h_s$ denote the analysis and synthesis in the auxiliary autoencoder. 
Further, $p_{\hat{\boldsymbol{y}} \mid \hat{\boldsymbol{z}}}(\hat{\boldsymbol{y}} \mid \hat{\boldsymbol{z}})$ is the estimated distribution conditioned on $\hat{\boldsymbol{z}}$. 
Like Ball\'{e} \etal, we use a normal distribution with a zero mean to extract spatial dependencies:

\begin{equation}
  p_{\hat{\boldsymbol{y}}|\hat{\boldsymbol{z}}}(\hat{\boldsymbol{y}}|\hat{\boldsymbol{z}})   \sim   N(\boldsymbol{0},\boldsymbol{\sigma^2})
  \label{eq:10}
\end{equation}

We use the RD-Loss (rate-distortion loss function) as:

\begin{equation}  \label{eq:11}
\mathcal{L}= \mathcal{R}(\hat{\boldsymbol{y}})+\mathcal{R}(\hat{\boldsymbol{z}})+\lambda \cdot \mathcal{D}(\boldsymbol{x}, \hat{\boldsymbol{x}})
\end{equation}

where $\lambda$ controls the rate-distortion trade-off and different $\lambda$ values corresponded to different bitrates. $\mathcal{R}(\hat{\boldsymbol{y}})$ and $\mathcal{R}(\hat{\boldsymbol{z}})$ denote the consumed bitrate in each of the bottlenecks. $\mathcal{D}(\boldsymbol{x}, \hat{\boldsymbol{x}})$ denotes the distortion term \cite{cheng2020learned}.


\begin{figure}[tb]
  \centering
   \includegraphics[trim=100 80 80 100, clip, width=.45\textwidth]{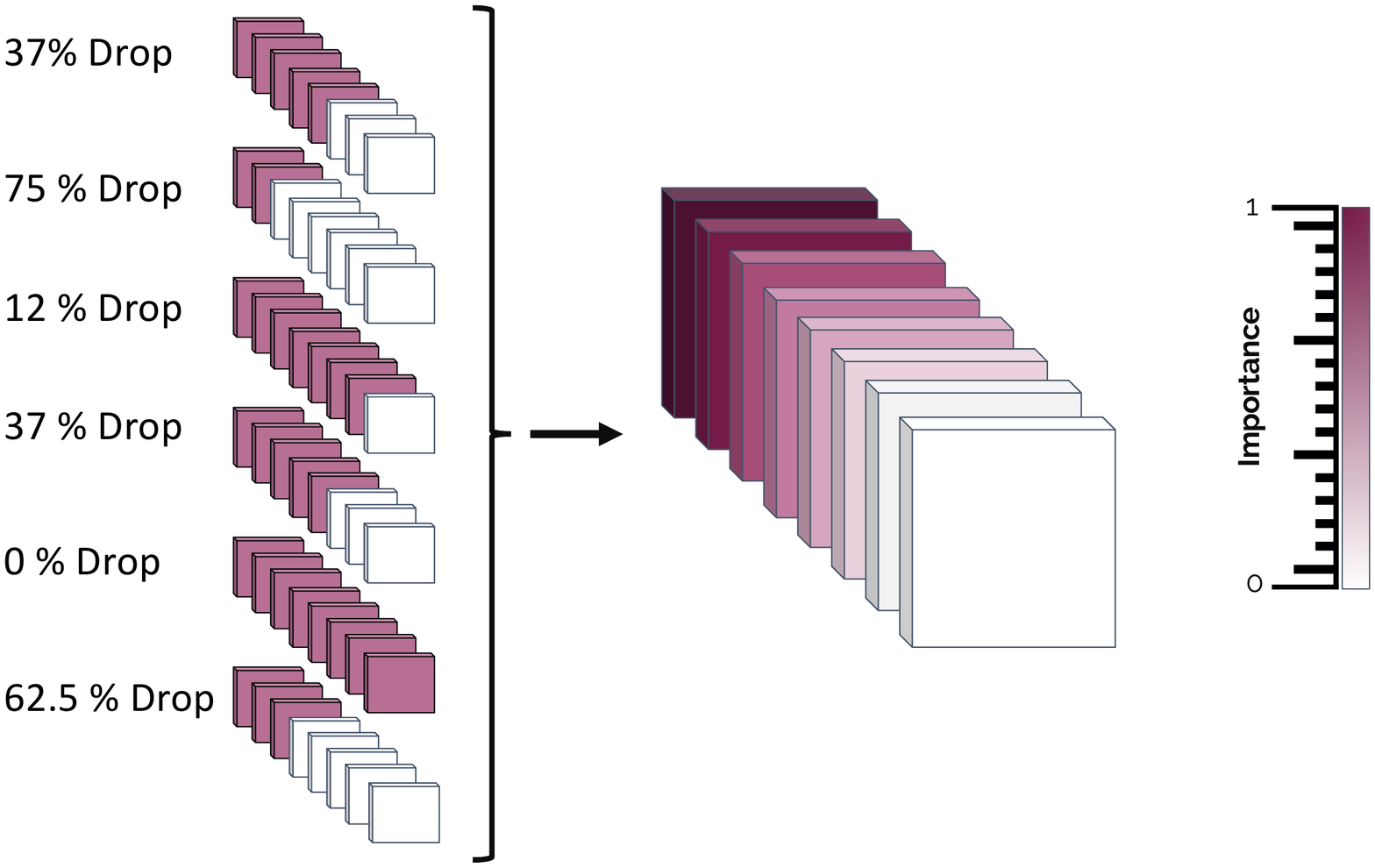}
   \caption{Tail-Drop Training: This image shows an example of the information distribution in the bottleneck $\mathcal{B}[8,M,M]$. If we train the model with the tail-drop, the training procedure puts the data in order of their importance. The colors show the importance of each filter.}
   \label{fig:Tail-Drop}
\end{figure}
 
\RestyleAlgo{ruled}
\SetKwComment{Comment}{/* }{ */}
\begin{algorithm}[tb]
    \caption{Double-tail-drop training. The red text marks our extensions over the original algorithm.}\label{alg:two}
    \KwData{Bottleneck of the latent: $\mathcal{B}_{lat}[C,M_l,M_l]$, \\Bottleneck of the hyperprior: $\mathcal{B}_{hp}[C,M_h,M_h]$, \\Number of batches: $N_{batch}$, \\Number of epochs: $N_{epoch}$, \\Batch size: $N_{batch-size}$}
    \KwResult{Progressively trained network}
    \label{alg:alg1}
    \For {$ 1 \leq e_{i} \leq N_{epoch}$}{
      \For {$ 1 \leq b_{i} \leq N_{batch}$}{
        \For {$ 1 \leq x_{i} \leq N_{batch-size}$}{
    
            $\mathcal{B}_{lat}$ = $ImageAnalysis$($\mathcal{D}[x_{i}]$)\;
            $\mathcal{B}_{hp}$ = $HyperAnalysis$($\mathcal{B}_{lat}$)\;
            
            \Comment {dropping from Tail}
            $K$ $\leftarrow$ Generate a sample from $\mathcal{U}(u_1,u_2)$\;
            \textcolor{red}{$\mathcal{B}_{hp}[K:C,M_h,M_h] = 0$}\;
            \textcolor{red}{$\mathcal{B}_{lat}[K:C,M_l,M_l] = 0$}\;
            
            $\mathcal{B}_{hp}$, $\mathcal{P}_{hp}$ = $HyperBottleneck$($\mathcal{B}_{lat}$)\;
    
            $\sigma$ = $HyperSynthesis$($\mathcal{B}_{hp}$)\;
            $\mathcal{B}_{lat}$, $\mathcal{P}_{lat}$=$ImageBottleneck$($\mathcal{B}_{lat}$, $\sigma$)\;
            
            \textcolor{red}{$\mathcal{P}_{hp}[K:C,M_h,M_h] = 1$}\;
            \textcolor{red}{$\mathcal{P}_{lat}[K:C,M_l,M_l] = 1$}\;
            
            $RecImage$ = $ImageAnalysis$($\mathcal{B}_{lat}$)\;
        }
      }
      $\mathcal{L}$ =  $RDLoss$($Rec$, $\mathcal{P}_{lat}$, $\mathcal{P}_{hp}$)\;
      Back-propagation\;
      Updating the model parameters $\theta$;
    }
\end{algorithm}


\subsection{Rateless autoencoder}\label{sec: Rateless autoencoder}


In \name, we adjust the training steps to store data in the bottleneck sorted by significance. 
Then by transmitting the filters of the bottleneck ordered by their significance, we inherently achieve progressive compression.
We achieve this goal by modifying the training steps of the model to enforce it to store the data in the bottleneck sorted by priority. 
We build on the tail-drop technique \cite{koike2020stochastic}, extend it, and, as a case study, apply it to the training of both the latent and hyper latent bottlenecks of the Ball\'{e} model \cite{balle2018variational}. We call this approach double-tail-drop (DTD).

We define the bottleneck $\mathcal{B}$ with $C$ filters of size $M*M$ as $\mathcal{B}_{[C, M, M]}$. 
Further, $\mathcal{B}_{[0:K, M, M]}$ denotes that we keep the first $K$ filters of the bottleneck and drop the others.
For simplicity, we define the loss function of an autoencoder with $K$ filter in the bottleneck as:
\begin{equation}
  \mathcal{L}(\theta,\phi,K)
\label{eq:15}
\end{equation}

We define our loss function as the following multi-objective optimization problem:

\begin{equation}
  \forall_{K\in \{0,1,2,...,N\}} \min  \mathcal{L}(\theta,\phi,K)
\label{eq:16}
\end{equation}

A naïve solution is weighted sum optimization to reduce the problem to a single objective function as follows:
\begin{equation}
  \min _{\theta, \phi} \sum_{K=1}^N \omega_L \mathcal{L}(\theta,\phi,K)
  \label{eq:important}
\end{equation}
Koike-Akino \etal \cite{koike2020stochastic} propose that stochastic tail-drop regularization during training can be interpreted as a weight $\omega_L$ for MLP neural networks. During training, for each batch, they draw a random number $K$ from the uniform distribution $\mathcal{U}(0,1)$ and then drop $K\%$ last filters from the bottleneck iteratively for $T$ times.
In other words, they calculate the loss $T$ times for each batch. The complexity of this training is in order of $\mathcal{O}(epoch*batch*T*\mathcal{C}_{loss-complexity})$. 
Their approach is specifically designed for MLP-based neural networks, is computationally heavy, and is not directly applicable to multi-stage autoencoders with auxiliary autoencoders and RD-Loss \cite{Balle2016end, balle2018variational, minnen2018joint}.
In the next section, we close this gap. 

\subsection{\name} \label{sec: Training procedure}
In this section, we introduce our double-tail-drop algorithm and how we integrate it into multi-stage autoencoders \cite{Balle2016end, balle2018variational, minnen2018joint}. 
Then, we discuss the progressive behavior of our model and explain how we extend the RD-Loss function to support double-tail-drop.

\begin{figure*}[tb]
  \centering
   \includegraphics[trim=10 160 30 160, clip, width=1.0\textwidth]{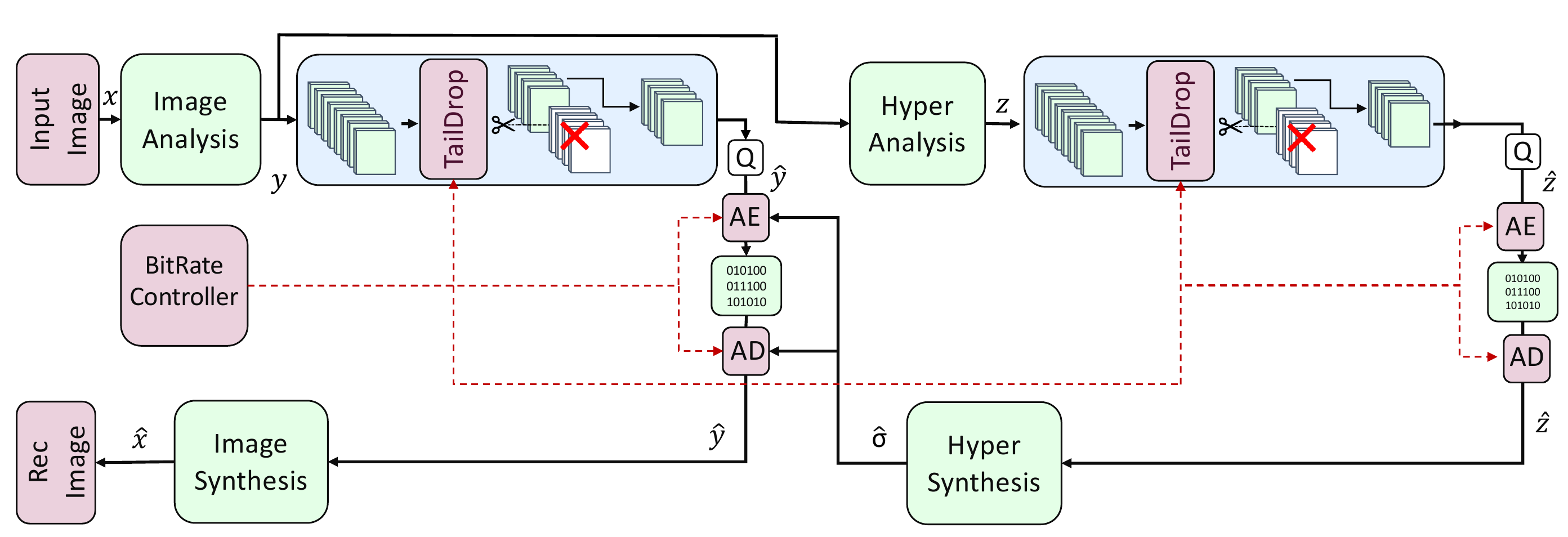}
   \caption{The network architecture of the hyperprior model \cite{balle2018variational} with double-tail-drop. Quantization is represented by Q, and arithmetic encoding and decoding are denoted by AE and AD, respectively. With the bitrate controller, we can determine the bitrate that we want to have.}
   \label{fig:overview}
\end{figure*}

\subsubsection{Double-tail-drop}
We extend the concept of tail-drop \cite{koike2020stochastic} to enable progressive image compression.
During training, for each input image, we draw a random number $K$ uniformly from $\mathcal{U}(u_1,u_2)$ and drop $K\%$ of the tail of the bottleneck for that specific image. 
If we do the tail dropping for each image of each batch, we distribute the knowledge like a CDF of uniform distributions $\mathcal{U}(u_1,u_2)$.
Our experimentation results show that the uniform distribution performs best among the common distributions. Furthermore, when we calculate the loss for each batch, we drop different percentages of each image in the batch. This allows us to optimize for all of the images in the batch simultaneously by calculating the gradient signals in a way that considers different tail drops to optimize the loss function (Eq. \ref{eq:16}).

Figure \ref{fig:Tail-Drop} shows an example of the tail-drop for a batch of size 6 and 8 filters in the bottleneck.  
With tail-drop training, we force the model to update gradient signals so that the first filters have more important information compared to the later filters. 
This is because the first filters are involved more frequently in the model training and dropped less often, whereas the final filters have less knowledge due to being dropped more frequently. The complexity of \name is in order of $\mathcal{O}(epoch*T*\mathcal{C}_{loss-complexity})$.

\subsubsection{Double-tail-drop for ImageAnalysis and HyperAnalysis networks}

\name is designed for CNN-based compression networks. 
To properly evaluate the benefits of \name, we employ the hyperprior model \cite{balle2018variational}, the fundamental architecture for most of today's learned image techniques.
Figure \ref{fig:overview} shows the position of the double-tail-drop in the Ball\'{e} (hyperprior) model. We employ two different tail-drop blocks: one for the latent and one for the hyperlatent.
As shown in Algorithm \ref{alg:alg1}, we drop a similar percentage in both tail-drop blocks. 
Therefore we only employ one random number generator. For example, if we keep 30 percent of the latent bottleneck $\mathcal{B}_{lat}$, we also keep 30 percent of the hyperlatent bottleneck $\mathcal{B}_{hp}$.
Nevertheless, we can apply tail-drop with different random number generators, but based on our results, the single generator performs better. 
The reason is that we need to train the model in a way that puts more important information in both bottlenecks simultaneously, which means if we want to send $K\%$ percent of both bottlenecks, it can select the most important parts of the hyperprior and the latent at the same time. 
Algorithm \ref{alg:alg1} shows the procedure of tail-drop training.
 
\subsubsection{Exploring progressive behavior}\label{sec:range}

\begin{figure*}
    \centering
    \begin{subfigure}[b]{0.495\textwidth}
        \centering
        \includegraphics[width=.95\textwidth]{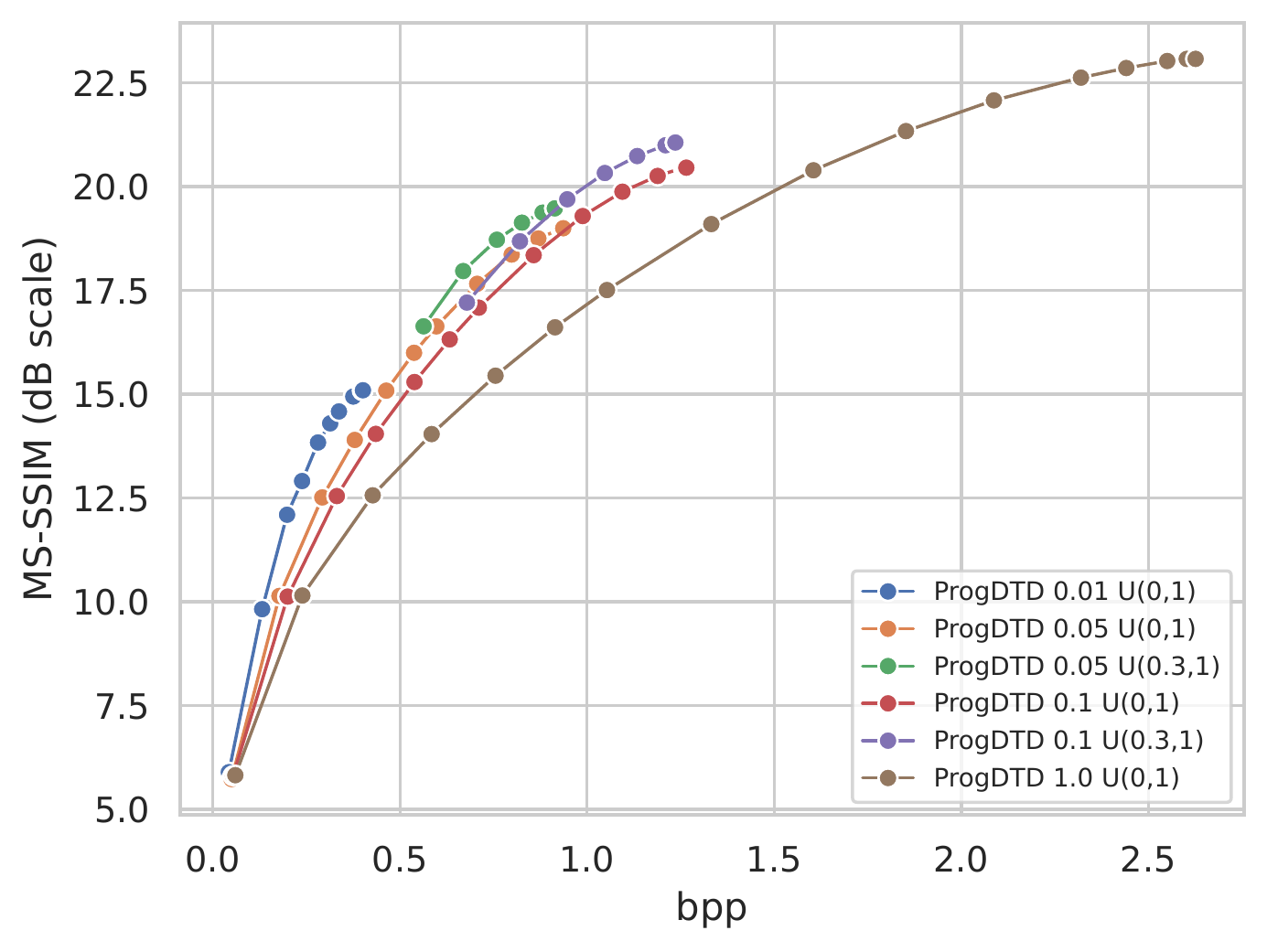}
        \caption[]%
        {{\small RD curves (MS-SSIM) of \name}}    
        \label{fig:DTD-MS-SSIM}
    \end{subfigure}
    \hfill
    \begin{subfigure}[b]{0.495\textwidth}
        \centering 
        \includegraphics[width=.95\textwidth]{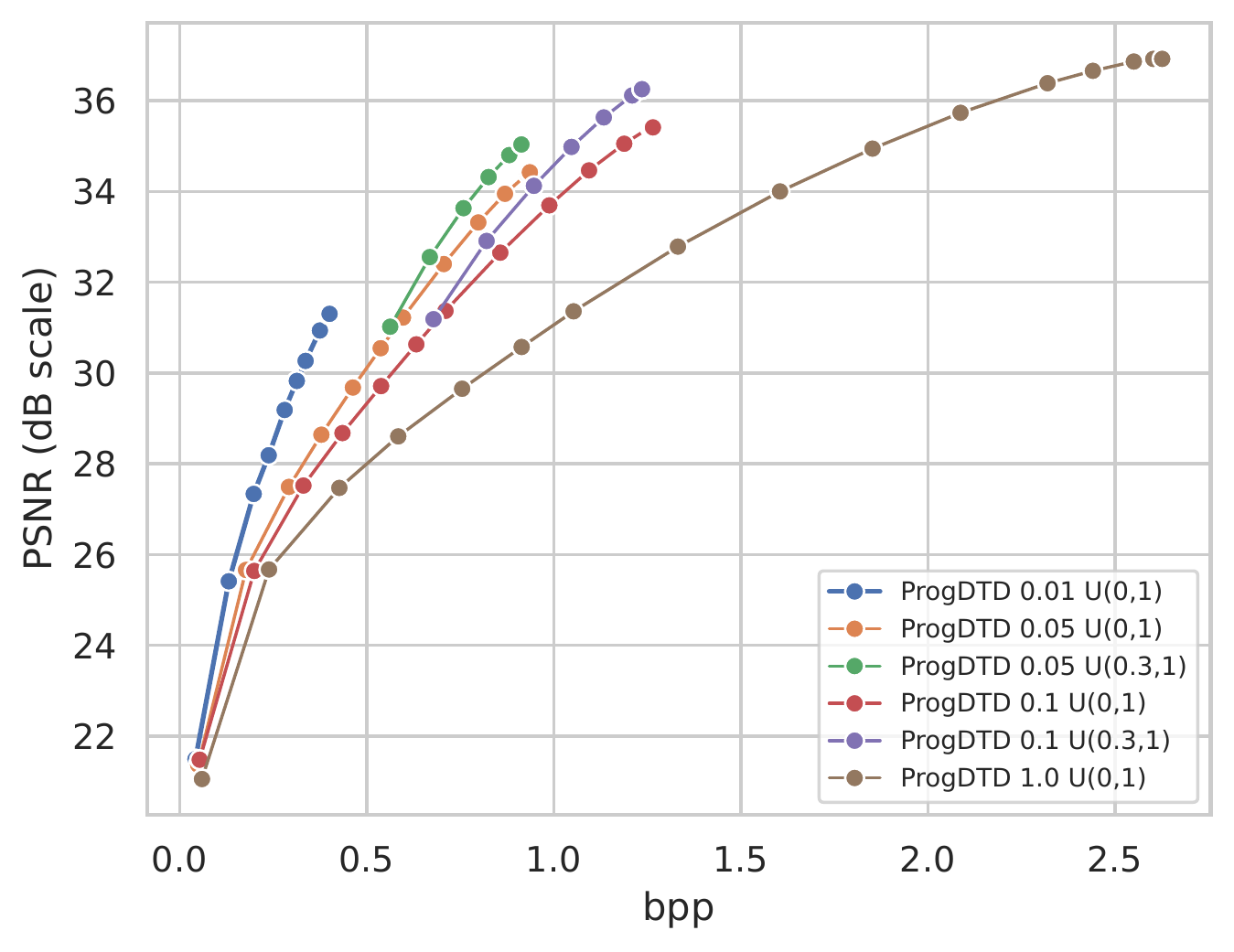}
        \caption[]%
        {{\small RD curves (PSNR) of \name}}    
        \label{fig:DTD-PSNR}
    \end{subfigure}
   \caption{RD performance (MS-SSIM and PSNR) comparison of the proposed \emph{\name} (trained with $\lambda=0.01, 0.05, 0.1, 1.0$ and also with $\mathcal{U}(0, 1)$ and $\mathcal{U}(0.3, 1)$ as the random number generator) on the KODAK dataset.}
\end{figure*}

As noted in the previous sections, we need to sample from $\mathcal{U}(u_1,u_2)$ during the training of \name, and by setting the range $(u_1,u_2)$, we specify the range of scalability. 
\name has the following limitation:
If we set a wide range of progressiveness, we lose some performance. For example, if we set $(u_1=0,u_2=1)$, our model can produce scalable results starting from less than 0.01 bpp. However, if we set $(u_1=0.3,u_2=1)$, we can achieve better performance with a narrower range of progressiveness. 
Thus, by considering the application of the compression model, we can specify the target bitrate range and train the model on that specific range.
In our evaluation in Section \ref{sec:result}, we further discuss and evaluate this effect, see Figures \ref{fig:DTD-ACC}, \ref{fig:DTD-MS-SSIM}, \ref{fig:DTD-PSNR}.

After training our model with \name, we send data in the bottlenecks according to its index.  
During training, we drop values by setting them to zero.
Further, on the decoder side, we know the positions of the missing filters and thus initialize these as zeros before feeding them with the received filters into the decoder. 
Overall, this technique leads to a significant reduction in the size of the compressed images.

As discussed in Section \ref{sec: Related Work}, most of the existing deep progressive models have limited steps in their scalability of progressiveness \cite{wallace1991jpeg, toderici2017full, cai2019novel, toderici2015variable}. 
In contrast, in our approach, we achieve $C\times (u_2-u_1)$ steps of scalability with $\mathcal{B}_{lat}[C,M,M]$ as a bottleneck and $\mathcal{U}(u_1,u_2)$ as a random number generator. 
For example, in our implementation, which we adapted from Ball\'{e} \cite{balle2018variational}, we have 196 progressive steps with $\mathcal{U}(u_1=0,u_2=1)$ and 129 progressive steps with $\mathcal{U}(u_1=0.3,u_2=1)$ (see Figures \ref{fig:DTD-ACC}, \ref{fig:DTD-MS-SSIM}, \ref{fig:DTD-PSNR}).


\begin{figure*}
    \centering
    \begin{subfigure}[b]{0.495\textwidth}
        \centering
        \includegraphics[width=.95\textwidth]{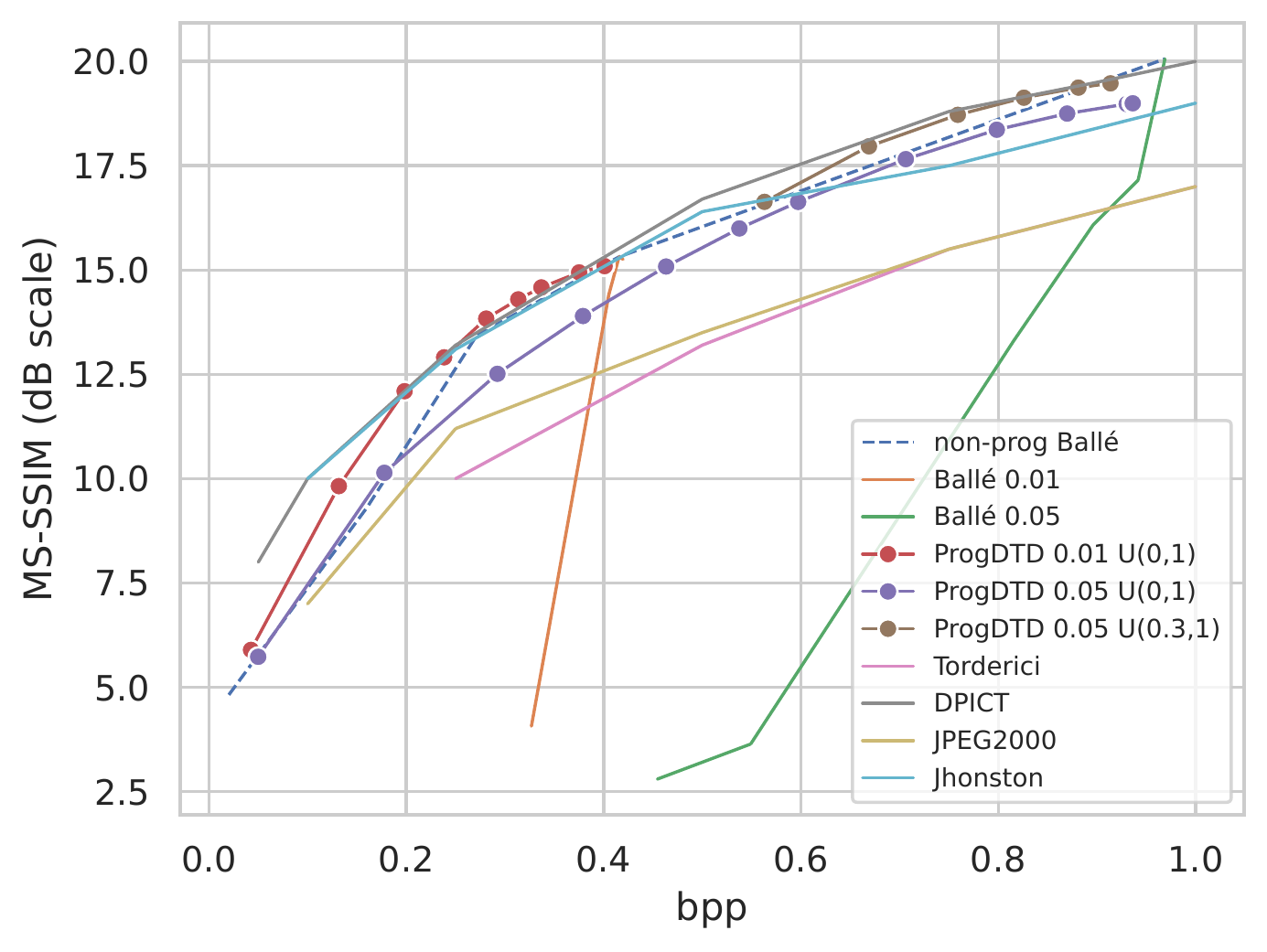}
        \caption[]%
        {{\small RD curves (MS-SSIM)}}    
        \label{fig:MS-SSIM}
    \end{subfigure}
    \hfill
    \begin{subfigure}[b]{0.495\textwidth}
        \centering 
        \includegraphics[width=.95\textwidth]{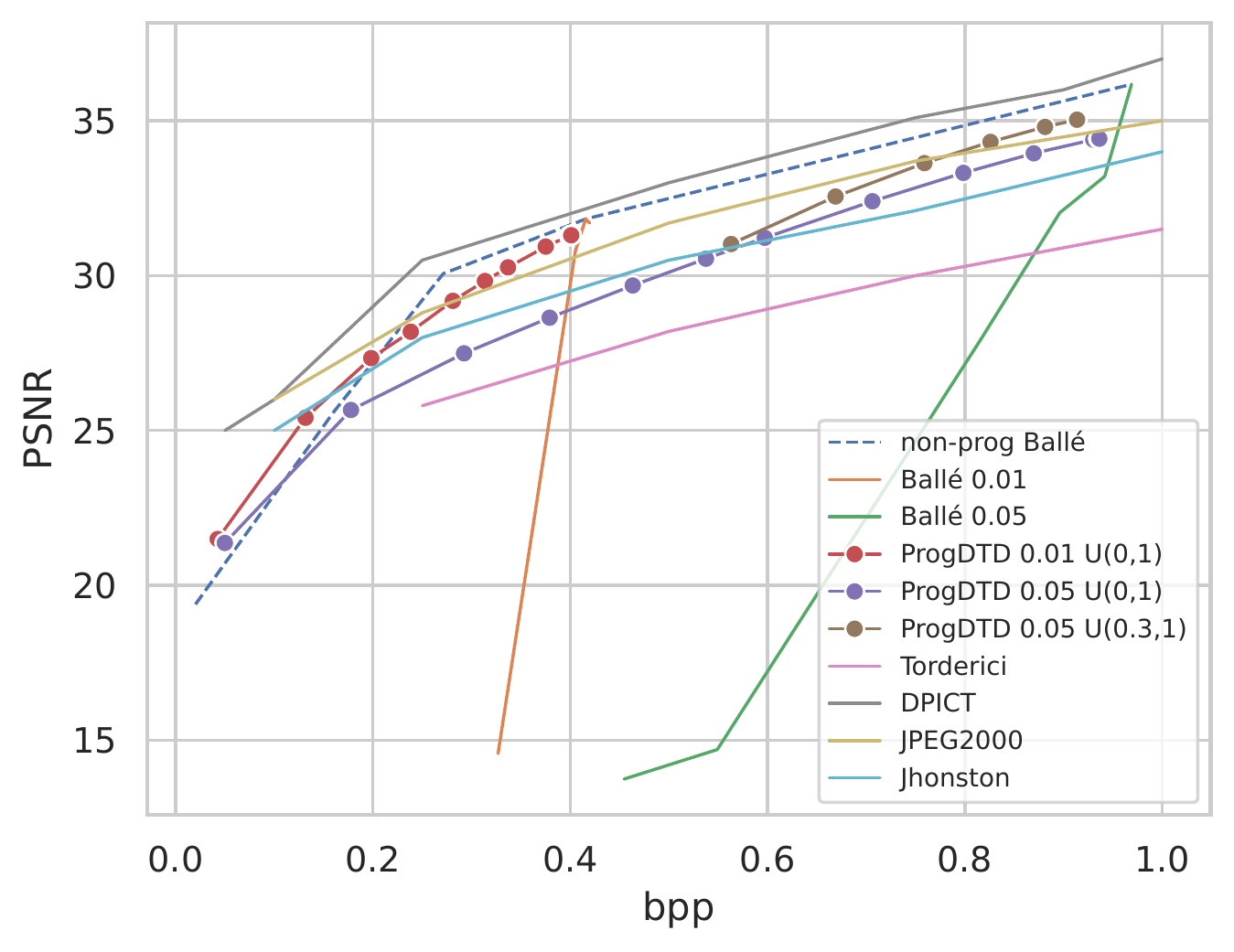}
        \caption[]%
        {{\small RD curves (PSNR)}}    
        \label{fig:PSNR}
    \end{subfigure}
   \caption{RD performance (MS-SSIM and PSNR) comparison of the proposed \emph{\name} (trained with $\lambda=0.01, 0.05$ and also with $\mathcal{U}(0, 1)$ and $\mathcal{U}(0.3, 1)$ as the random number generator),  \emph{non-progressive-Ball\'{e}} (trained with $\lambda=0.0001, 0.001, 0.005, 0.01, 0.05$) and \emph{standard-Ball\'{e}} (trained with $\lambda=0.01, 0.05$), DPICT (2022)\cite{lee2022dpict}, JPEG2000 \cite{skodras2001jpeg}, Johnston (2018) \etal \cite{johnston2018improved} and Torderici (2015) \etal \cite{toderici2015variable} on the KODAK dataset. Note that the \emph{non-progressive-Ball\'{e}} (dashed line) is a collection of Ball\'{e} models which have been trained with different $\lambda$, and it is \textbf{not} progressive.}
\end{figure*}

\begin{figure*}
    \centering
    \begin{subfigure}[b]{0.495\textwidth}
        \centering
        \includegraphics[width=.95\textwidth]{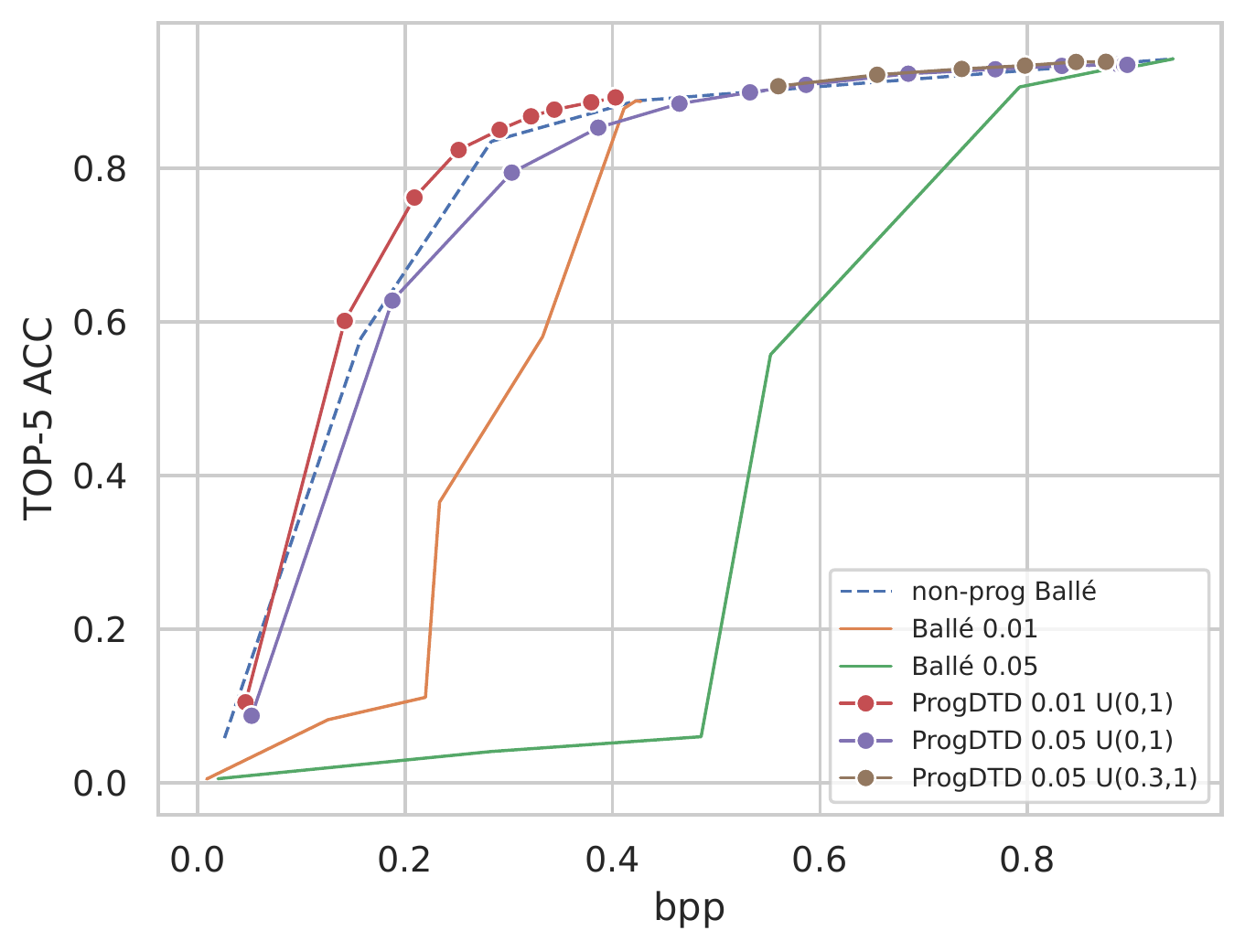}
        \caption[]%
        {{\small RD curves (ACC)}}    
        \label{fig:ACC}
    \end{subfigure}
    \hfill
    \begin{subfigure}[b]{0.495\textwidth}
        \centering 
        \includegraphics[width=.95\textwidth]{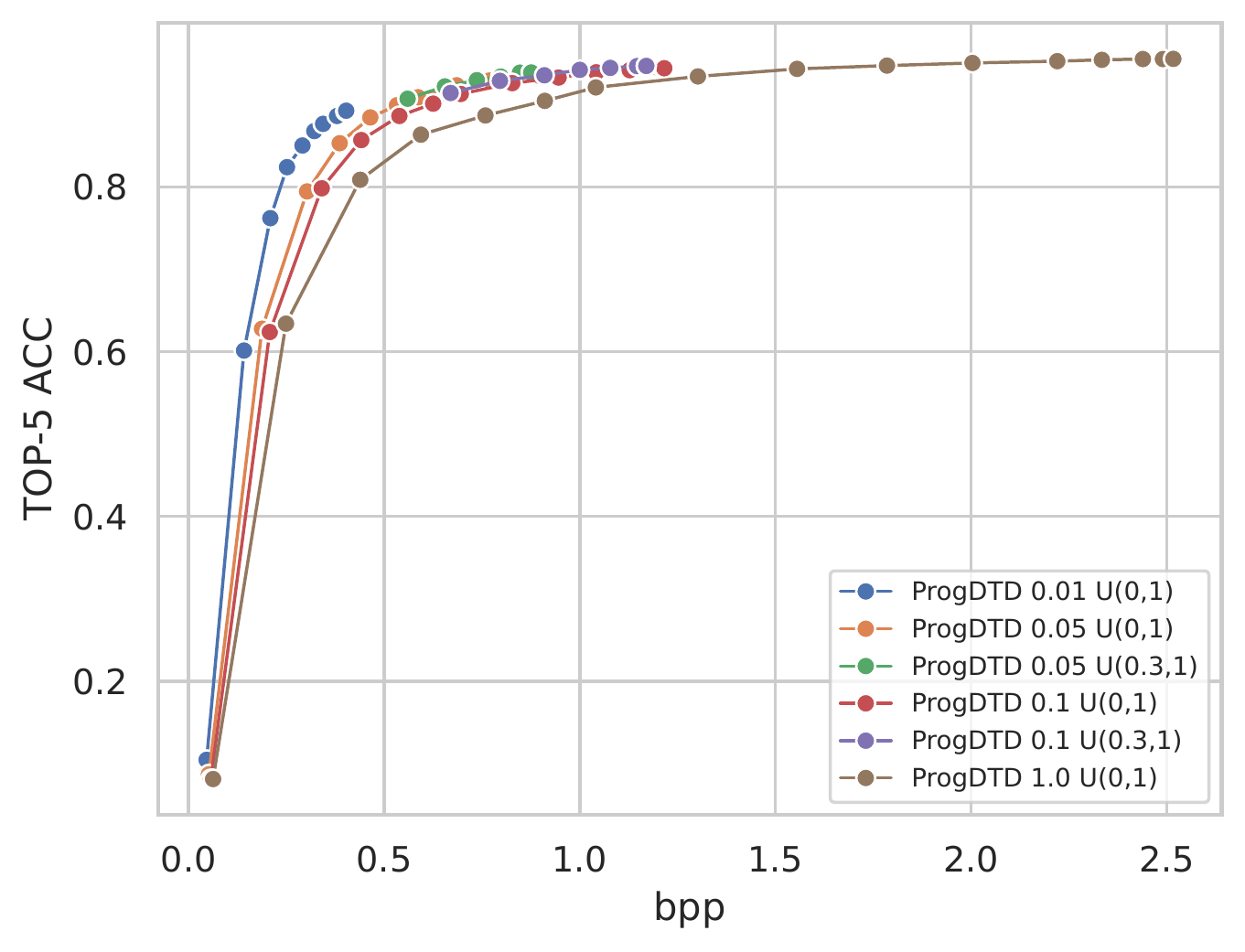}
        \caption[]%
        {{\small RD curves (ACC) of \name}}    
        \label{fig:DTD-ACC}
    \end{subfigure}
    \hfill

   \caption{RA (Rate-Accuracy) performance of \emph{\name} with same configuration like \ref{fig:MS-SSIM}, \ref{fig:PSNR}, \ref{fig:DTD-PSNR}, \ref{fig:DTD-MS-SSIM}. Note that the \emph{non-progressive-Ball\'{e}} (dashed line) is a collection of Ball\'{e} models which have been trained with different $\lambda$, and it is \textbf{not} progressive.}
\end{figure*}

\subsubsection{Rate Distortion loss function}

Since we do not send the dropped values to the decoder, we must change the bitrate calculator to consider this dropping during training. 
As shown in the Algorithm \ref{alg:alg1}, we remove the effect of dropped values by setting $\mathcal{P}_{lat}[K:C,M_h,M_h]$ and $\mathcal{P}_{hp}[K:C,M_h,M_h]$ to one.

Then we can rewrite Eq. \ref{eq:11} as:

\begin{equation}
\begin{split}
R+\lambda \cdot D=& \mathbb{E}_{\boldsymbol{x} \sim p_{\boldsymbol{x}}}\left[-\log _2 p\prime_{\hat{\boldsymbol{y}}}(\hat{\boldsymbol{y}})\right] \\+&
\mathbb{E}_{\boldsymbol{x} \sim p_{\boldsymbol{x}}}\left[-\log _2 p\prime_{\hat{\boldsymbol{z}}}(\hat{\boldsymbol{z}})\right]\\+&\lambda \cdot \mathbb{E}_{\boldsymbol{x} \sim p_{\boldsymbol{x}}}\|\boldsymbol{x}-\hat{\boldsymbol{x}}\|_2^2
\end{split}
\end{equation}

where $p\prime_{\hat{\boldsymbol{z}}}(\hat{\boldsymbol{z}})$ and $p\prime_{\hat{\boldsymbol{y}}}(\hat{\boldsymbol{y}})$ are a modified version of $p_{\hat{\boldsymbol{z}}}(\hat{\boldsymbol{z}})$ and $p_{\hat{\boldsymbol{y}}}(\hat{\boldsymbol{y}})$.

\section{Evaluation}
\label{sec:result}

In this section, we evaluate the performance of \name. 
We begin with a brief discussion of implementation details and datasets. Next, we present parameters before focusing on the performance evaluation. 
We conclude this section with a discussion of the results. 

\subsection{Implementation and Datasets}
For the model architecture, we build on the Ball\'{e} model \cite{balle2018variational} and integrate \name into the implementation of the Ball\'{e} model by Facebook Research \cite{muckley2021neuralcompression}.
As datasets, we use the Vimeo90k dataset \cite{xue2019video} for training, and both the KODAK~\cite{kodak} and ImageNet \cite{deng2009imagenet} datasets for evaluation. 
For training, we extract 30,000 images with $256\times256$ patches from the Vimeo dataset, and we use the Adam \cite{kingma2014adam} optimizer with a batch size of 4 and a learning rate of 0.0001. 
We train the model for 150 epochs using MSE as the loss function. 
As typical, we evaluate \name on the KODAK dataset, which consists of 24 images of resolution $512\times768$. 
We resize the images to $512\times512$ and then extract four patches with a size of $256\times256$ each. Finally, we calculate the bitrates by bits per pixel and evaluate our model using PSNR and MS-SSIM as metrics. 
As image compression is often combined with other tasks such as classification, we -- in addition -- evaluate the classification accuracy of EfficientNet-B0 \cite{tan2019efficientnet} on the ImageNet \cite{deng2009imagenet} dataset, which gets the reconstructed images from \name as inputs.

\subsection{Evaluation Setup}
To evaluate our approach, we train our model with  $\lambda=0.01, 0.05$ and the Ball\'{e} model with $\lambda=0.0001, 0.001, 0.005, 0.01, 0.05$ and utilize $\mathcal{U}(0, 1)$ and $\mathcal{U}(0.3, 1)$ as the ranges for random number generation in our filters. First, we compare \name with \emph{non-progressive-Ball\'{e}}, trained for different lambdas, to show that \name achieves comparable performance across different bitrates. 
Second, since \name provides progressive image compression, we compare our model and the standard Ball\'{e} model with the same scenario of sending each filter by its index. 
We call this \emph{{standard-Ball\'{e}}} and show that it fails to achieve good performance when reducing the bitrate.

\subsection{RD-curves with MS-SSIM and PSNR}

As discussed in the design section, \name has a trade-off that needs to be evaluated: 
Prioritizing a broader range of progressiveness will come at the cost of losing some performance. 
Conversely, if we limit the range of progressiveness, we achieve better performance. 
Figure \ref{fig:MS-SSIM} presents MS-SSIM and shows that ${\name^{\lambda=0.05}_{\mathcal{U}(0, 1)}}$, which has been trained with $\lambda=0.05$ and $\mathcal{U}(0, 1)$, has a performance closed to \emph{non-progressive-Ball\'{e}}. 
In other words, ${\name^{\lambda=0.05}_{\mathcal{U}(0, 1)}}$ provides progressive compression and achieves a compression quality with a slight drop in MS-SSIM when compared to the standard Ball\'{e} which has been trained separately for different $\lambda$. 
We can compensate for this decline in performance by limiting the range of the random number generator $\mathcal{U}(u_1, u_2)$ and thereby limiting the degree of progressiveness. 
For example, Figure \ref{fig:MS-SSIM} shows that ${\name^{\lambda=0.05}_{\mathcal{U}(0.3, 1)}}$ reaches the same performance as \emph{non-progressive-Ball\'{e}}.

Next, we evaluate PSNR. 
Figure \ref{fig:PSNR} shows that \name can not reach the performance of \emph{non-progressive-Ball\'{e}} in terms of PSNR.
This is due to the following reason: 
As \name utilizes the tail-drop approach for training, the last filters of the bottleneck are less involved in the training and therefore receive less gradient to optimize.
This leads to the consequence that we can not use the full capacity of the bottleneck. 
Since PSNR focuses on the actual values, and we do not use the full capacity of the bottleneck, this explains the decline in PSNR when compared to the performance \emph{non-progressive-Ball\'{e}},  

As discussed in Section \ref{sec: Proposed Algorithm}, \name achieves progressive compression by sending filters based on their index. 
In Figures \ref{fig:MS-SSIM} and \ref{fig:PSNR} we compare our model with \emph{standard-Ball\'{e}} (trained separately for $\lambda=0.01 , 0.05$), which shows the result of the standard Ball\'{e} model when we drop filters based on their index. 
Our results show that -- unlike \emph{standard-Ball\'{e}} -- our model has a graceful decrease in terms of MS-SSIM and PSNR and outperforms \emph{standard-Ball\'{e}}, showing the effect of double-tail-drop training. 
To appropriately evaluate the performance of \name, we compare it further to selected state-of-the-art models like DPICT (2022)\cite{lee2022dpict}, JPEG2000 \cite{skodras2001jpeg}, Johnston (2018) \etal \cite{johnston2018improved} and Torderici (2015) \etal \cite{toderici2015variable}. 
Figure \ref{fig:MS-SSIM} shows that our results are comparable to or better than those of the aforementioned models.

To visually evaluate the reconstruction performance of \name and compare it to state of the art, we feed one sample from the KODAK dataset to \emph{\name}, \emph{non-progressive-Ball\'{e}} and \emph{standard-Ball\'{e}}.
Figure \ref{fig:Plots} plots the reconstructed image for multiple compression levels.
It shows that \emph{\name} and \emph{non-progressive-Ball\'{e}} have a similar reconstruction result, meanwhile, the \emph{standard-Ball\'{e}} model does not have any good reconstruction as soon as we drop from the bottleneck. 

\subsection{RD-curve with accuracy}
One of the use cases of image and video compression is for AI tasks. 
For example, suppose we have an edge device that does not have the required hardware capability to run a classification task but can compress and send data to a powerful server over a network connection with variable bandwidth \cite{yao2020deep, ko2018edge, li2018jalad, yao2018fastdeepiot}.
In such a case of variable bandwidth (e.g., cellular connectivity) and especially when classification tasks have a deadline, we argue that progressive image compression provides substantial benefits. 

In this scenario, we care about classification accuracy instead of reconstruction metrics like PSNR and MS-SSIM, which we discussed before.
We feed the reconstructed image to EfficientNet-B0 \cite{tan2019efficientnet} and calculate the top-5 accuracy to evaluate this. 
Figure \ref{fig:ACC} shows that ${\name^{\lambda=0.01}_{\mathcal{U}(0, 1)}}$ and ${\name^{\lambda=0.05}_{\mathcal{U}(0.3, 1)}}$ with the smaller scalability ranges outperform \emph{non-progressive-Ball\'{e}}, while ${\name^{\lambda=0.05}_{\mathcal{U}(0, 1)}}$ with the larger scalability range from 0 bpp to 1 bpp, has on par performance compared to \emph{non-progressive-Ball\'{e}}. Moreover, \name outperforms \emph{standard-Ball\'{e}}  since \emph{standard-Ball\'{e}} does not deal well with dropping parts of the bottleneck.

\subsection{Discussion}
\label{sec:discussion}

Overall, our evaluation results show that \name achieves a good and often even a comparable performance compared to its non-progressive counterpart. 
However, despite its benefits, the design of \name is limited.
We train the model by utilizing tail-drop, and by dropping from the tail of the bottleneck, we are forcing the model to put the most important information in the first filters.
Because of this, the last filters in the bottleneck are less involved in the model training, and they will get less gradient to optimize. 
In other words, with \name, we do not use the full capacity of the bottleneck, as shown in Figures \ref{fig:DTD-PSNR}, \ref{fig:DTD-MS-SSIM}, and \ref{fig:DTD-ACC}.
If we train our model with a high $\lambda$, we will get an increased range of scalability, but at the same time, we lose some performance in the smaller bitrates.
In contrast, if we train with a smaller $\lambda$, we achieve better performance at the same bpp, but we have limited our progressiveness.

\subsection{Efficiency of \name}

As we mentioned, \name is a training method that transforms non-progressive compression models to progressive ones and does not add any parameters or complexity to the base model.
Therefore, the encoding and decoding times remain the same. 
Like other learned progressive compression models \cite{lee2022dpict, toderici2015variable}, we only add decoding time by iterating the decoder when we want to load an image progressively. However, when receiving incomplete data due to time-critical constraints, by running the decoder only once after the cut-off time, we can have a preview of the image even with a small portion of the data, which would not work for a non-progressive model, as shown in Figures \ref{fig:MS-SSIM}, \ref{fig:PSNR},\ref{fig:ACC}.

\section{Conclusions}
\label{sec:conclusion}
This paper presents \name, a training approach that transforms non-progressive image compression methods into progressive ones. 
The idea behind \name is based on the fact that information stored within the bottleneck of a compression model can have varying levels of importance. 
\name modifies the training steps to prioritize data storage in the bottleneck based on its importance and transmits the data based on their level of importance to achieve progressive compression. 
This technique is designed for CNN-based learned image compression models and requires no additional parameters. 
We use the hyperprior model, a fundamental structure of learned image compression methods, to evaluate the effectiveness of \name. Our experimental results show that \name performs comparably to its non-progressive counterparts and other state-of-the-art progressive models in terms of MS-SSIM and accuracy.
For our future work, we will explore the performance of \name on other learned compression models to verify that \name works with any CNN-based model.

\section*{Acknowledgements}
\label{sec:acknowledgements}
This project has received funding from the 
Federal Ministry for Digital and Transport under the CAPTN-F\"{o}rde 5G project grant no.~45FGU139\_H
and
Federal Ministry for Economic Affairs and Climate Action under the Marispace-X project grant no.~68GX21002E.

{\small
\bibliographystyle{ieee_fullname}
\bibliography{ProgDTD}

\begin{thebibliography}{10}\itemsep=-1pt

\bibitem{agustsson2019generative}
Eirikur Agustsson, Michael Tschannen, Fabian Mentzer, Radu Timofte, and Luc~Van
  Gool.
\newblock Generative adversarial networks for extreme learned image
  compression.
\newblock In {\em Proceedings of the IEEE/CVF International Conference on
  Computer Vision}, pages 221--231, 2019.

\bibitem{Balle2015density}
Johannes Ball{\'e}, Valero Laparra, and Eero~P Simoncelli.
\newblock Density modeling of images using a generalized normalization
  transformation.
\newblock {\em arXiv preprint arXiv:1511.06281}, 2015.

\bibitem{Balle2016end}
Johannes Ball{\'e}, Valero Laparra, and Eero~P Simoncelli.
\newblock End-to-end optimized image compression.
\newblock {\em arXiv preprint arXiv:1611.01704}, 2016.

\bibitem{balle2018variational}
Johannes Ball{\'e}, David Minnen, Saurabh Singh, Sung~Jin Hwang, and Nick
  Johnston.
\newblock Variational image compression with a scale hyperprior.
\newblock {\em arXiv preprint arXiv:1802.01436}, 2018.

\bibitem{bpg}
Bpg image format.
\newblock \url{https://bellard.org/bpg/}.
\newblock Accessed: 2023-02-14.

\bibitem{cai2018efficient}
Chunlei Cai, Li Chen, Xiaoyun Zhang, and Zhiyong Gao.
\newblock Efficient variable rate image compression with multi-scale
  decomposition network.
\newblock {\em IEEE Transactions on Circuits and Systems for Video Technology},
  29(12):3687--3700, 2018.

\bibitem{cai2019novel}
Chunlei Cai, Li Chen, Xiaoyun Zhang, Guo Lu, and Zhiyong Gao.
\newblock A novel deep progressive image compression framework.
\newblock In {\em 2019 Picture Coding Symposium (PCS)}, pages 1--5. IEEE, 2019.

\bibitem{cheng2020learned}
Zhengxue Cheng, Heming Sun, Masaru Takeuchi, and Jiro Katto.
\newblock Learned image compression with discretized gaussian mixture
  likelihoods and attention modules.
\newblock In {\em Proceedings of the IEEE/CVF Conference on Computer Vision and
  Pattern Recognition}, pages 7939--7948, 2020.

\bibitem{choi2019variable}
Yoojin Choi, Mostafa El-Khamy, and Jungwon Lee.
\newblock Variable rate deep image compression with a conditional autoencoder.
\newblock In {\em Proceedings of the IEEE/CVF International Conference on
  Computer Vision}, pages 3146--3154, 2019.

\bibitem{cui2021asymmetric}
Ze Cui, Jing Wang, Shangyin Gao, Tiansheng Guo, Yihui Feng, and Bo Bai.
\newblock Asymmetric gained deep image compression with continuous rate
  adaptation.
\newblock In {\em Proceedings of the IEEE/CVF Conference on Computer Vision and
  Pattern Recognition}, pages 10532--10541, 2021.

\bibitem{deng2009imagenet}
Jia Deng, Wei Dong, Richard Socher, Li-Jia Li, Kai Li, and Li Fei-Fei.
\newblock Imagenet: A large-scale hierarchical image database.
\newblock In {\em 2009 IEEE conference on computer vision and pattern
  recognition}, pages 248--255. Ieee, 2009.

\bibitem{diao2020drasic}
Enmao Diao, Jie Ding, and Vahid Tarokh.
\newblock Drasic: Distributed recurrent autoencoder for scalable image
  compression.
\newblock In {\em 2020 Data Compression Conference (DCC)}, pages 3--12. IEEE,
  2020.

\bibitem{goodfellow2020generative}
Ian Goodfellow, Jean Pouget-Abadie, Mehdi Mirza, Bing Xu, David Warde-Farley,
  Sherjil Ozair, Aaron Courville, and Yoshua Bengio.
\newblock Generative adversarial networks.
\newblock {\em Communications of the ACM}, 63(11):139--144, 2020.

\bibitem{gregor2016towards}
Karol Gregor, Frederic Besse, Danilo Jimenez~Rezende, Ivo Danihelka, and Daan
  Wierstra.
\newblock Towards conceptual compression.
\newblock {\em Advances In Neural Information Processing Systems}, 29, 2016.

\bibitem{hochreiter1997long}
Sepp Hochreiter and J{\"u}rgen Schmidhuber.
\newblock Long short-term memory.
\newblock {\em Neural computation}, 9(8):1735--1780, 1997.

\bibitem{johnston2018improved}
Nick Johnston, Damien Vincent, David Minnen, Michele Covell, Saurabh Singh,
  Troy Chinen, Sung~Jin Hwang, Joel Shor, and George Toderici.
\newblock Improved lossy image compression with priming and spatially adaptive
  bit rates for recurrent networks.
\newblock In {\em Proceedings of the IEEE Conference on Computer Vision and
  Pattern Recognition}, pages 4385--4393, 2018.

\bibitem{kingma2014adam}
Diederik~P Kingma and Jimmy Ba.
\newblock Adam: A method for stochastic optimization.
\newblock {\em arXiv preprint arXiv:1412.6980}, 2014.

\bibitem{ko2018edge}
Jong~Hwan Ko, Taesik Na, Mohammad~Faisal Amir, and Saibal Mukhopadhyay.
\newblock Edge-host partitioning of deep neural networks with feature space
  encoding for resource-constrained internet-of-things platforms.
\newblock In {\em 2018 15th IEEE International Conference on Advanced Video and
  Signal Based Surveillance (AVSS)}, pages 1--6. IEEE, 2018.

\bibitem{kodak}
Eastman kodak (1993). kodak lossless true color image suite (photocd pcd0992).
\newblock \url{https://r0k.us/graphics/kodak}.

\bibitem{koike2020stochastic}
Toshiaki Koike-Akino and Ye Wang.
\newblock Stochastic bottleneck: Rateless auto-encoder for flexible
  dimensionality reduction.
\newblock In {\em 2020 IEEE International Symposium on Information Theory
  (ISIT)}, pages 2735--2740. IEEE, 2020.

\bibitem{krizhevsky2017imagenet}
Alex Krizhevsky, Ilya Sutskever, and Geoffrey~E Hinton.
\newblock Imagenet classification with deep convolutional neural networks.
\newblock {\em Communications of the ACM}, 60(6):84--90, 2017.

\bibitem{langdon1981compression}
Glen Langdon and Jorma Rissanen.
\newblock Compression of black-white images with arithmetic coding.
\newblock {\em IEEE Transactions on Communications}, 29(6):858--867, 1981.

\bibitem{lee2018context}
Jooyoung Lee, Seunghyun Cho, and Seung-Kwon Beack.
\newblock Context-adaptive entropy model for end-to-end optimized image
  compression.
\newblock {\em arXiv preprint arXiv:1809.10452}, 2018.

\bibitem{lee2022dpict}
Jae-Han Lee, Seungmin Jeon, Kwang~Pyo Choi, Youngo Park, and Chang-Su Kim.
\newblock Dpict: Deep progressive image compression using trit-planes.
\newblock In {\em Proceedings of the IEEE/CVF Conference on Computer Vision and
  Pattern Recognition}, pages 16113--16122, 2022.

\bibitem{li2018jalad}
Hongshan Li, Chenghao Hu, Jingyan Jiang, Zhi Wang, Yonggang Wen, and Wenwu Zhu.
\newblock Jalad: Joint accuracy-and latency-aware deep structure decoupling for
  edge-cloud execution.
\newblock In {\em 2018 IEEE 24th international conference on parallel and
  distributed systems (ICPADS)}, pages 671--678. IEEE, 2018.

\bibitem{minnen2018joint}
David Minnen, Johannes Ball{\'e}, and George~D Toderici.
\newblock Joint autoregressive and hierarchical priors for learned image
  compression.
\newblock {\em Advances in neural information processing systems}, 31, 2018.

\bibitem{muckley2021neuralcompression}
Matthew Muckley, Jordan Juravsky, Daniel Severo, Mannat Singh, Quentin Duval,
  and Karen Ullrich.
\newblock Neuralcompression.
\newblock \url{https://github.com/facebookresearch/NeuralCompression}, 2021.

\bibitem{nakanishi2019neural}
Ken~M Nakanishi, Shin-ichi Maeda, Takeru Miyato, and Daisuke Okanohara.
\newblock Neural multi-scale image compression.
\newblock In {\em Computer Vision--ACCV 2018: 14th Asian Conference on Computer
  Vision, Perth, Australia, December 2--6, 2018, Revised Selected Papers, Part
  VI 14}, pages 718--732. Springer, 2019.

\bibitem{ohm2005advances}
J-R Ohm.
\newblock Advances in scalable video coding.
\newblock {\em Proceedings of the IEEE}, 93(1):42--56, 2005.

\bibitem{rippel2017real}
Oren Rippel and Lubomir Bourdev.
\newblock Real-time adaptive image compression.
\newblock In {\em International Conference on Machine Learning}, pages
  2922--2930. PMLR, 2017.

\bibitem{scholz2008nonlinear}
Matthias Scholz, Martin Fraunholz, and Joachim Selbig.
\newblock Nonlinear principal component analysis: neural network models and
  applications.
\newblock In {\em Principal manifolds for data visualization and dimension
  reduction}, pages 44--67. Springer, 2008.

\bibitem{skodras2001jpeg}
Athanassios Skodras, Charilaos Christopoulos, and Touradj Ebrahimi.
\newblock The jpeg 2000 still image compression standard.
\newblock {\em IEEE Signal processing magazine}, 18(5):36--58, 2001.

\bibitem{su2020scalable}
Rige Su, Zhengxue Cheng, Heming Sun, and Jiro Katto.
\newblock Scalable learned image compression with a recurrent neural
  networks-based hyperprior.
\newblock In {\em 2020 IEEE International Conference on Image Processing
  (ICIP)}, pages 3369--3373. IEEE, 2020.

\bibitem{tan2019efficientnet}
Mingxing Tan and Quoc Le.
\newblock Efficientnet: Rethinking model scaling for convolutional neural
  networks.
\newblock In {\em International conference on machine learning}, pages
  6105--6114. PMLR, 2019.

\bibitem{theis2017lossy}
Lucas Theis, Wenzhe Shi, Andrew Cunningham, and Ferenc Husz{\'a}r.
\newblock Lossy image compression with compressive autoencoders.
\newblock {\em arXiv preprint arXiv:1703.00395}, 2017.

\bibitem{toderici2015variable}
George Toderici, Sean~M O'Malley, Sung~Jin Hwang, Damien Vincent, David Minnen,
  Shumeet Baluja, Michele Covell, and Rahul Sukthankar.
\newblock Variable rate image compression with recurrent neural networks.
\newblock {\em arXiv preprint arXiv:1511.06085}, 2015.

\bibitem{toderici2017full}
George Toderici, Damien Vincent, Nick Johnston, Sung Jin~Hwang, David Minnen,
  Joel Shor, and Michele Covell.
\newblock Full resolution image compression with recurrent neural networks.
\newblock In {\em Proceedings of the IEEE conference on Computer Vision and
  Pattern Recognition}, pages 5306--5314, 2017.

\bibitem{tschannen2018deep}
Michael Tschannen, Eirikur Agustsson, and Mario Lucic.
\newblock Deep generative models for distribution-preserving lossy compression.
\newblock {\em Advances in neural information processing systems}, 31, 2018.

\bibitem{van2016pixel}
A{\"a}ron Van Den~Oord, Nal Kalchbrenner, and Koray Kavukcuoglu.
\newblock Pixel recurrent neural networks.
\newblock In {\em International conference on machine learning}, pages
  1747--1756. PMLR, 2016.

\bibitem{wallace1991jpeg}
Gregory~K Wallace.
\newblock The jpeg still picture compression standard.
\newblock {\em Communications of the ACM}, 34(4):30--44, 1991.

\bibitem{wallace1992jpeg}
Gregory~K Wallace.
\newblock The jpeg still picture compression standard.
\newblock {\em IEEE transactions on consumer electronics}, 38(1):xviii--xxxiv,
  1992.

\bibitem{webp}
Webp.
\newblock \url{https://developers.google.com/speed/webp/docs/compression}.
\newblock Accessed: 2023-02-14.

\bibitem{wold1987principal}
Svante Wold, Kim Esbensen, and Paul Geladi.
\newblock Principal component analysis.
\newblock {\em Chemometrics and intelligent laboratory systems}, 2(1-3):37--52,
  1987.

\bibitem{xue2019video}
Tianfan Xue, Baian Chen, Jiajun Wu, Donglai Wei, and William~T Freeman.
\newblock Video enhancement with task-oriented flow.
\newblock {\em International Journal of Computer Vision}, 127:1106--1125, 2019.

\bibitem{yang2021slimmable}
Fei Yang, Luis Herranz, Yongmei Cheng, and Mikhail~G Mozerov.
\newblock Slimmable compressive autoencoders for practical neural image
  compression.
\newblock In {\em Proceedings of the IEEE/CVF Conference on Computer Vision and
  Pattern Recognition}, pages 4998--5007, 2021.

\bibitem{yang2020variable}
Fei Yang, Luis Herranz, Joost Van De~Weijer, Jos{\'e} A~Iglesias Guiti{\'a}n,
  Antonio~M L{\'o}pez, and Mikhail~G Mozerov.
\newblock Variable rate deep image compression with modulated autoencoder.
\newblock {\em IEEE Signal Processing Letters}, 27:331--335, 2020.

\bibitem{yao2020deep}
Shuochao Yao, Jinyang Li, Dongxin Liu, Tianshi Wang, Shengzhong Liu, Huajie
  Shao, and Tarek Abdelzaher.
\newblock Deep compressive offloading: Speeding up neural network inference by
  trading edge computation for network latency.
\newblock In {\em Proceedings of the 18th Conference on Embedded Networked
  Sensor Systems}, pages 476--488, 2020.

\bibitem{yao2018fastdeepiot}
Shuochao Yao, Yiran Zhao, Huajie Shao, ShengZhong Liu, Dongxin Liu, Lu Su, and
  Tarek Abdelzaher.
\newblock Fastdeepiot: Towards understanding and optimizing neural network
  execution time on mobile and embedded devices.
\newblock In {\em Proceedings of the 16th ACM Conference on Embedded Networked
  Sensor Systems}, pages 278--291, 2018.

\end{thebibliography}
}

\end{document}